\documentclass[sigconf,nonacm]{acmart}
\AtBeginDocument{%
  }
\usepackage{balance}
\usepackage{xcolor}
\usepackage{colortbl}
\usepackage{enumitem}
\definecolor{tabcol}{HTML}{a2d2ff}
\author{Hanxiao Wang}
\affiliation{%
 \institution{CASIA, KAUST}
}

\author{Biao Zhang}
\affiliation{%
 \institution{KAUST}
}
\author{Jonathan Klein}
\affiliation{%
 \institution{KAUST}
}
\author{Dominik L. Michels}
\affiliation{%
 \institution{KAUST}
}

\author{Dongming Yan}
\affiliation{%
 \institution{CASIA}
}
\author{Peter Wonka}
\affiliation{%
 \institution{KAUST}
}

\begin{document}

\title{Autoregressive Generation of Static and Growing Trees}

\begin{abstract}
We propose a transformer architecture and training strategy for tree generation. The architecture processes data at multiple resolutions and has an hourglass shape, with middle layers processing fewer tokens than outer layers. Similar to convolutional networks, we introduce longer-range skip connections to complement this multi-resolution approach. The key advantages of this architecture are the faster processing speed and lower memory consumption. We are, therefore, able to process more complex trees than would be possible with a vanilla transformer architecture. Furthermore, we extend this approach to perform image-to-tree and point-cloud-to-tree conditional generation and to simulate the tree growth processes, generating 4D trees. Empirical results validate our approach in terms of speed, memory consumption, and generation quality.
\end{abstract}





\begin{teaserfigure}
    \centering
    \includegraphics[width=0.95\textwidth, height=7.5cm]{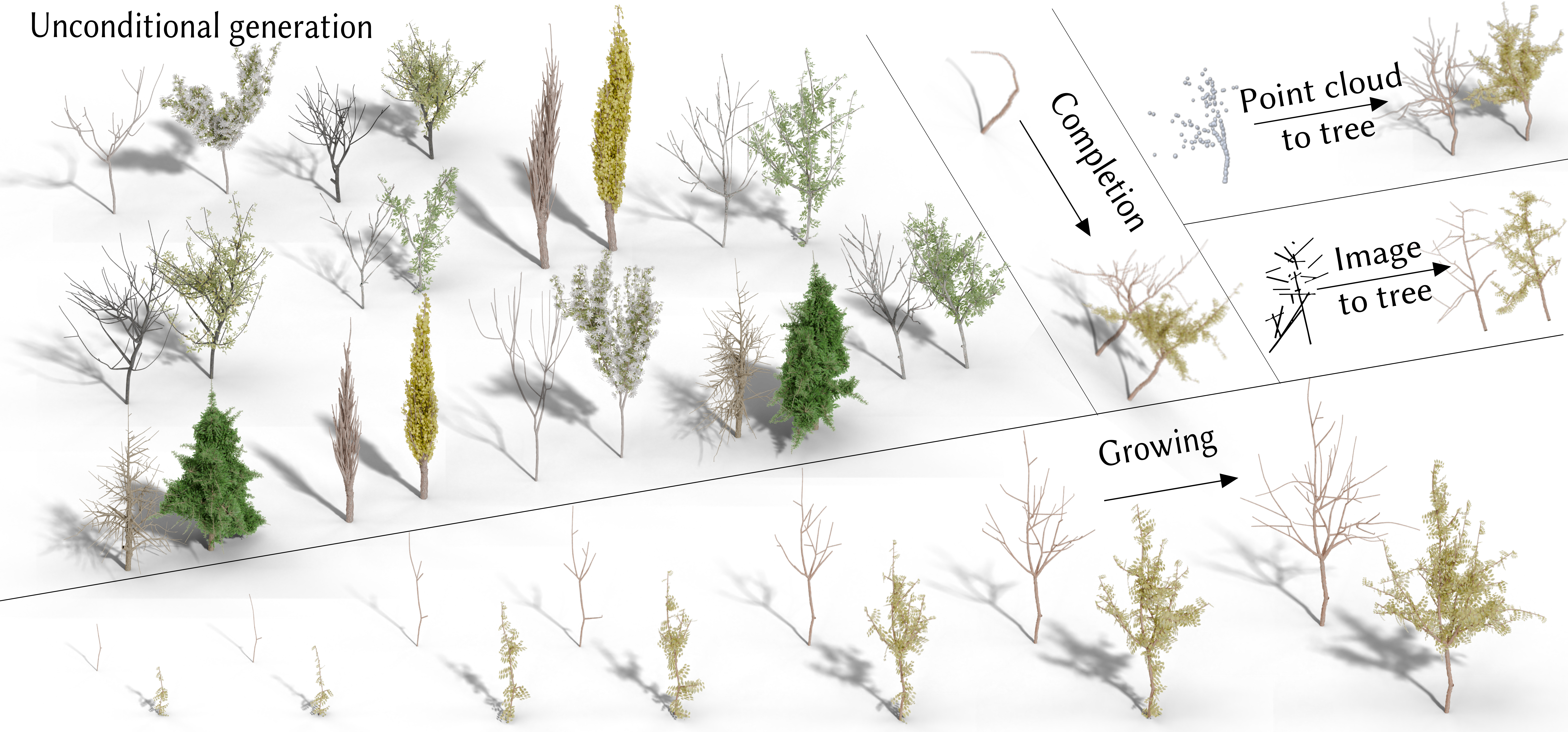}
    \vspace{-15pt}
    \caption{We introduce a data structure, transformer architecture, and training strategy for the autoregressive generation of trees. We show various applications such as the unconditional generation of static trees, the unconditional generation of growing trees, tree completion, point cloud to tree conversion, and 2D line sketch to tree conversion. Our method generates the tree skeleton while leaves are added procedurally. }
    \label{fig:teaser}
\end{teaserfigure}
\maketitle
\section{Introduction}

Tree generation is a classical problem in computer graphics and simulation. Trees typically exhibit complex hierarchical structures and diverse branching patterns. Traditional tree generation methods are mainly based on procedural techniques or rule-based simulations, which employ predefined growth rules or biological principles to produce plausible tree structures \cite{smith1984plants, lindenmayer1968mathematical,  reeves1985approximate,  de1988plant, lewis1999three}. Although these approaches can be effective in specific settings, they require extensive parameter tuning and lack flexibility in adapting to different tree species and environments. Consequently, researchers continue to look for adaptive and scalable solutions capable of generating high-quality trees more efficiently.

The rise of deep generative models has opened up new avenues for generating complex structures. Notable examples include generative adversarial networks (GANs) \cite{10.1145/3422622}, variational autoencoders (VAEs) \cite{kingma2013auto}, autoregressive models \cite{esser2021taming}, and diffusion models \cite{ho2020denoising} which have demonstrated remarkable abilities in various modalities, such as images, text, point clouds, and videos \cite{luo2021diffusion, vaswani2017attention,hao2024meshtron, chen2024meshanything2, ho2022video}. However, directly extending these deep generative methods to tree generation poses unique challenges. Trees possess hierarchical and recursive structures, in which branching patterns must maintain spatial coherence while also following biological growth principles. Traditional deep generative methods, though highly effective for flat or non-recursive data, often face difficulties in maintaining structural consistency, preserving branch integrity, and achieving efficient training when applied to tree structures \cite{ zhou2023deeptree, lee2025tree, li2024svdtree, guo2020inverse}. Therefore, a specialized approach remains necessary to fully harness the capabilities of deep generative models while respecting the core constraints of tree generation.

To tackle these challenges of tree generation, we propose a novel deep generative model that balances hierarchical structure preservation with computational efficiency. Our method introduces several key innovations that substantially improve the realism, coherence, and scalability of generated tree models.

\begin{enumerate}[leftmargin=*]
\item The first deep generative approach explicitly tailored to tree structures.  
   We present the first deep generative solution specifically designed for tree-structured data. By accounting for the hierarchical dependencies inherent in trees, our model provides greater flexibility and higher-quality tree generation compared to prior methods.

\item Streamlined parameterization and representation.  
   We leverage an efficient parameterization strategy based on preorder traversal, which ensures hierarchical coherence while reducing parameter dimensionality. This design preserves structural integrity and improves computational efficiency, enabling faster training and inference even for large and complex trees.

\item Hourglass transformer for multi-resolution processing and improved training efficiency.  
   We implement an \textit{hourglass-shaped} transformer architecture \cite{nawrot2021hierarchical, hao2024meshtron}, wherein the intermediate layers handle fewer tokens than the outer layers. We incorporate skip connections to complement the multi-resolution design. It yields faster processing speed and lower memory consumption, enabling us to handle more complex trees than conventional transformer architectures.

\item Extensions for conditional generation and 4D trees.  
   Beyond unconditional generation, our framework supports image-to-tree and point-cloud-to-tree conditional generation and 4D tree growth simulations. The latter enables the dynamic evolution of trees while maintaining structural coherence. Empirical results confirm improvements in speed, memory usage, and overall generation quality across these tasks.
\end{enumerate}


\section{Related Work}
\noindent\textbf{Early tree modeling.}
Early studies employed fractals and repetitive structures~\cite{smith1984plants}, L-systems~\cite{lindenmayer1968mathematical,prusinkiewicz1986graphical, prusinkiewicz2012algorithmic}, particle systems~\cite{reeves1985approximate, neubert2007approximate}, and biological growth rules \cite{de1988plant}. 
The influence of physical and environmental conditions on the tree structure has also been taken into account, considering biomechanical properties \cite{Zhao2013,10.1145/3072959.3073655,Maggioli23}, fire~\cite{Pirk:2017}, wind~\cite{Pirk14ToG,Shao:2021:GraphLearning}, and climate \cite{10.1145/3528223.3530146}. The modeling of root systems has also been addressed for a variety of trees \cite{Li:2023:Rhizomorph}. On another trajectory, modeling large ecosystems \cite{makowski2019synthetic} has been addressed, as well as the simulation of devastating wildfires \cite{Haedrich:2021:Wildfires,Kokosza:2024:Scintilla} within these systems.
A related problem is inverse procedural modeling, which aims to encode given inputs as a procedural model~\cite{stava2014inverse,guo2020inverse}.

\noindent\textbf{Tree modeling using deep learning.}
Advancements in deep learning have facilitated significant progress in tree generation, ranging from tree reconstruction from images \cite{Li:2021:ReconstructBotanicalTrees} to the use of transformer architectures to generate L-system grammars~\cite{lee2023latent}. As relying on the syntax of the L-system restricts precise control over intricate details in generated structures, Zhou et al.~\cite{zhou2023deeptree} explore deep learning for tree generation by focusing on the local context. However, their approach mainly considers the parent nodes and lacks comprehensive global structural awareness. Lee et al.~\cite{lee2025tree} introduce a dataset and employ diffusion models to generate trees from a single image. Similarly, Li et al.~\cite{li2024svdtree} utilize voxel-based diffusion to produce coarse semantic representations for tree reconstructions from single-view images. 



These deep learning-based methods generally depend on indirect representations, such as L-systems or voxel-based diffusion techniques, which limit their ability to exert fine-grained control over tree structures. In contrast, our approach seeks to bridge this gap by investigating a native representation of trees, thereby facilitating more effective and detailed control over the generated models.

\noindent\textbf{3D generative modeling.}
Generative models have made significant advancements, evolving from variational autoencoders (VAEs) \cite{kingma2013auto} and generative adversarial networks (GANs)~\cite{10.1145/3422622} to diffusion models~\cite{song2020score, ho2020denoising, Na:2024:LennardJonesLayer} and autoregressive models~\cite{esser2021taming}. 
In 3D data generation, early approaches utilized voxel-based representations~\cite{voxelhane2017hierarchical}, which suffer from high memory consumption and limited resolution. Point cloud-based methods~\cite{luo2021diffusion} are more efficient but lack explicit surface connectivity. Implicit field-based models~\cite{chen2019learning, zhang20233dshape2vecset,zhang2024lagem} represent 3D shapes as continuous functions, enabling high-resolution surface reconstruction but often requiring complex post-processing to extract meshes. Mesh-based generative models aim to generate 3D meshes with explicit surface representations directly~\cite{chen2020bsp,alliegro2023polydiff}. 

Autoregressive transformer-based methods such as MeshGPT \cite{siddiqui2024meshgpt}, LLaMA-Mesh~\cite{wang2024llama}, MeshXL~\cite{chen2024meshxl}, PivotMesh~\cite{weng2024pivotmesh}, and others have been proposed to improve mesh generation by predicting mesh elements using self-attention. MeshAnything ~\cite{chen2024meshanything} and MeshAnythingV2 ~\cite{chen2024meshanything2}, and EdgeRunner~\cite{tang2024edgerunner} implement mesh generation conditioned on point clouds. However, when performing unconditional generation, most of these methods can only generate meshes with 800 to 1600 faces, which is insufficient for modeling complex structures like trees.
Our approach addresses this limitation by proposing a representation more suitable for tree modeling. Leveraging the properties of this representation, we propose to use an hourglass transformer architecture~\cite{nawrot2021hierarchical,hao2024meshtron} to achieve higher efficiency and generate detailed meshes with significantly more faces, capturing the intricate structures of trees more effectively.

\section{Method}

\begin{figure}[t]
  \centering
  \includegraphics[width=\linewidth]{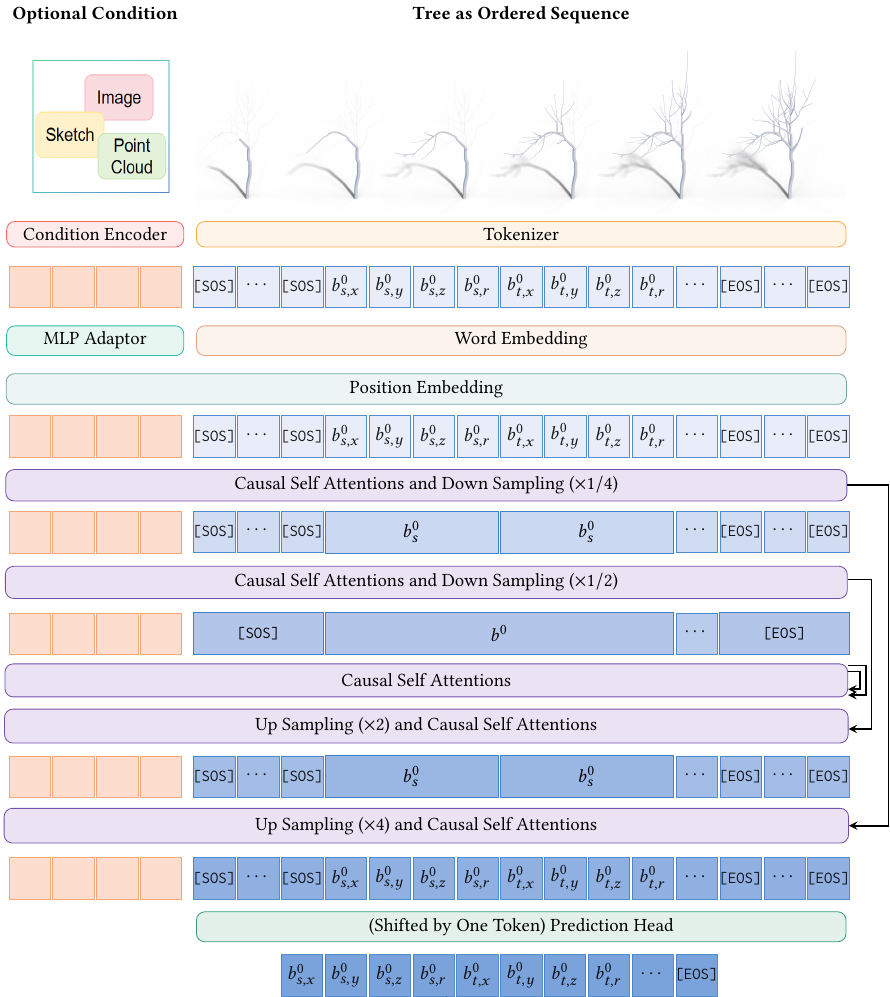}
  \caption{Architecture of our HourglassTree method}
  \vspace{-10pt}
  \Description{random trees}
  \label{backbone}
  \vskip -0.2cm
\end{figure}

This section presents our model, \textit{HourglassTree}, which comprises a carefully designed tree representation and an Hourglass Transformer architecture. Section~\ref{sec:representation} introduces our tree representation and the tokenization method we employ. Section~\ref{sec:hourglass_transformer} details the Hourglass~\cite{nawrot2021hierarchical,hao2024meshtron} Transformer architecture used for efficient learning.

\subsection{Tree Representation}
\label{sec:representation}

\subsubsection{Parameterization}

Traditional methods for tree generation, including procedural and rule-based approaches, as well as growth simulations, often produce meshes with a very large number of faces, sometimes exceeding hundreds of thousands. This complexity poses challenges for face-based generation methods based on deep learning that typically handle meshes with considerably fewer faces.

To address this issue, we propose a more concise parameterization of trees by viewing a tree as a set of branches, which aligns closely with the natural characteristics of real-world trees. Each branch is represented as two end-points with associated radii, forming a cylinder in 3D space (see~Fig.~\ref{fig:data_structure}). Specifically, we define:
\begin{figure}[!ht]
  \centering
  \includegraphics[width=0.6\linewidth]{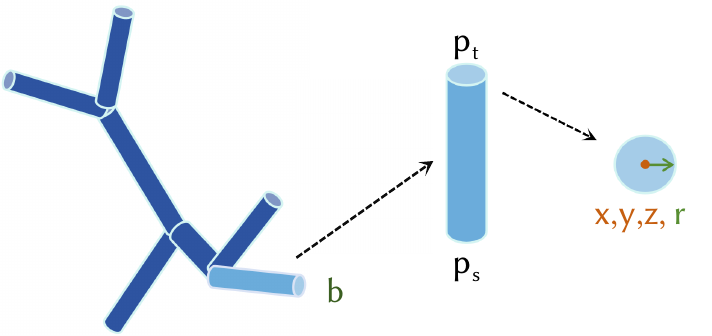}
  \caption{An illustration of our data structure.}
  \label{fig:data_structure}
  \vspace{-5pt}
  \vskip -0.2cm
\end{figure}
\begin{itemize}[leftmargin=*]
    \item \textbf{Tree}: $\boldsymbol{\tau} = \{\mathbf{b}^0, \mathbf{b}^1, \dots, \mathbf{b}^{n-1}\}$, a set of $n$ branches composing the entire tree structure. Note that $\boldsymbol{\tau}$ is an unordered set.
    \item \textbf{Branch}: $\mathbf{b} = (\mathbf{p}_s, \mathbf{p}_t)$ (or using the notation $ \mathbf{b}^i= (\mathbf{b}_s^i, \mathbf{b}_t^i)$), a pair of vertices defining the branch's two end-points.
    \item \textbf{Point}: $\mathbf{p} = (x, y, z, r)$, where $(x, y, z)$ denotes the 3D coordinates and $r$ represents the radius at that vertex. We also use the notation $ \mathbf{b}^i= ((\mathbf{b}_{s,x}^i, \mathbf{b}_{s,y}^i, \mathbf{b}_{s,z}^i, \mathbf{b}_{s,r}^i), (\mathbf{b}_{t,x}^i, \mathbf{b}_{t,y}^i, \mathbf{b}_{t,z}^i, \mathbf{b}_{t,r}^i))$.
\end{itemize}


With this parameterization, a tree consisting of 200 branches requires only 1,600 parameters (each branch has 2 points, and each point has 4 parameters), significantly reducing the complexity compared to directly generating a mesh with over 100,000 parameters.

\subsubsection{Tokenization}
\begin{figure}[t]
\vspace{-10pt}
  \centering
  \includegraphics[width=\linewidth]{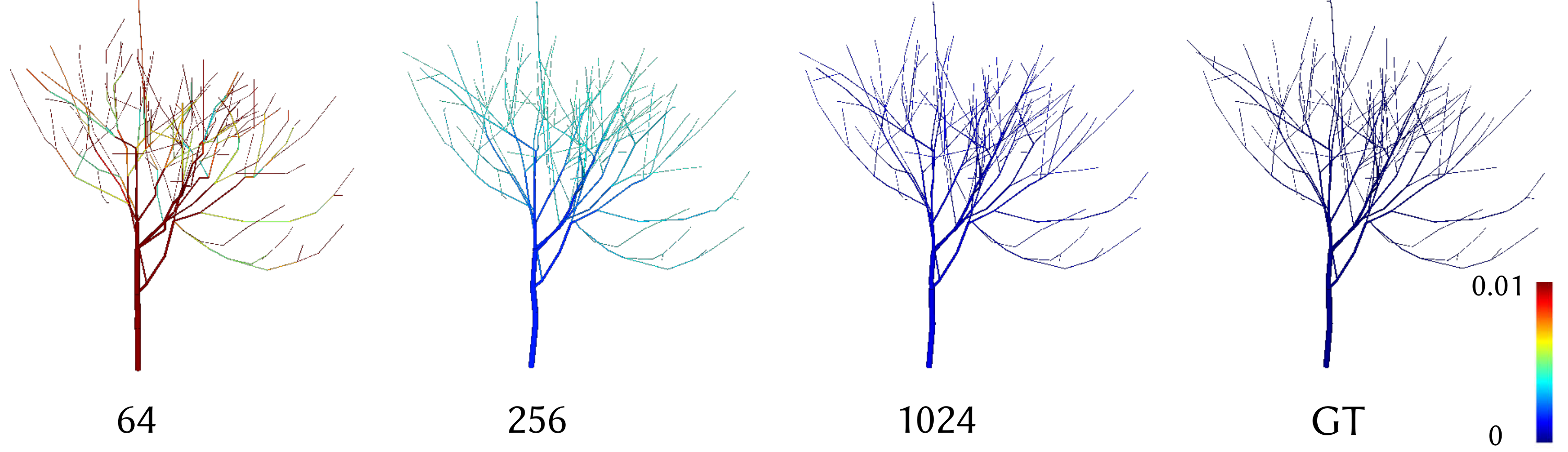}
  \vskip -0.2cm
  \caption{Different resolutions of data quantization.}
    \vspace{-5pt}
  \label{fig:data_quantization}
\end{figure}

To utilize an autoregressive transformer model, we discretize the continuous parameters into discrete tokens. To balance complexity and quality, each continuous value is quantized into one of 256 bins (\textit{e.g.} 256 distinct labels). The spatial coordinates \((x, y, z)\) and the radius \(r\) use different bin boundary values due to their different statistical distributions, as shown in Fig.~\ref{fig:data_quantization}. 
We introduce special tokens for sequence control, including \texttt{[SOS]}, \texttt{[EOS]}, and \texttt{[PAD]} for sequence initialization, termination, and padding, respectively. For training efficiency, we prepend 8 \texttt{[SOS]} tokens and append 8 \texttt{[EOS]} tokens. For example, for elm trees, the sequence is padded to a fixed length of $8 \times 202 = 1,616$ tokens, sufficient to represent a tree with 200 branches.

\subsubsection{Ordering of Branches}

The order of tokens is crucial for training autoregressive transformers. Some existing approaches sort mesh elements based on spatial coordinates or employ specifically designed ordering strategies. In our setting, we define an ordering for the branches that captures the structural dependencies inherent in tree geometry.

\begin{figure}[!ht]
  \centering
      \vspace{-5pt}
  \includegraphics[width=\linewidth]{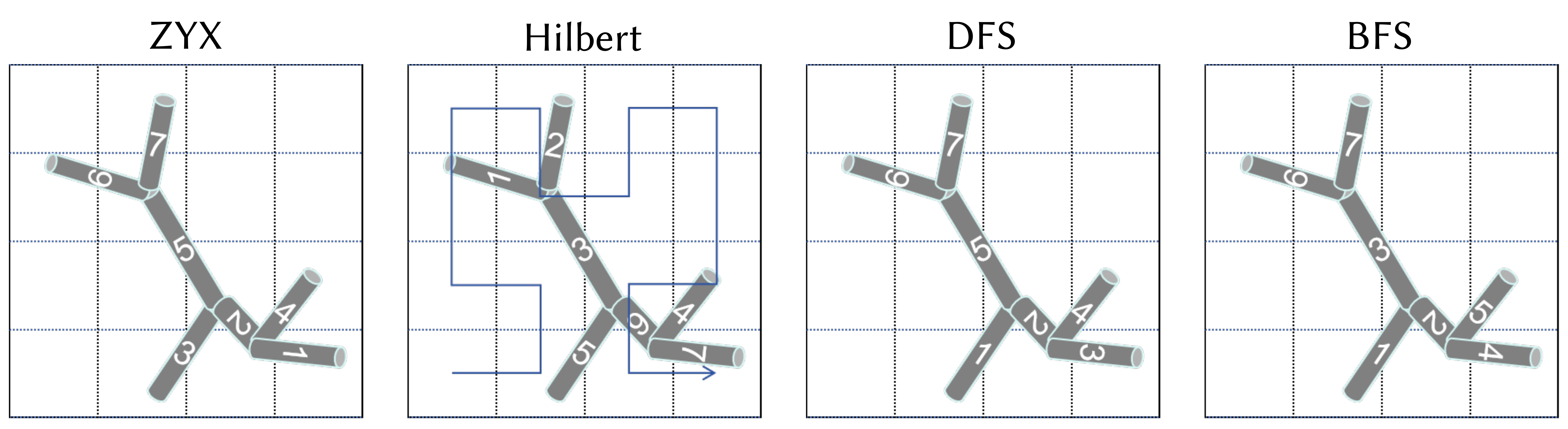}
  \caption{Different strategies for ordering tokens.}
  \label{order_illu}
  \Description{Comparison of token orders in tree generation.}
    \vspace{-10pt}
\end{figure}

A popular approach is to sort branches by their zyx order (sorting by z then by y then by x coordinates descendingly)~\cite{nash2020polygen}, as shown in Figure~\ref{order_illu}. This is also observed in the image context, where the simplest ordering tends to be the best~\cite{esser2021taming}. In point cloud learning, OctFormer and PTv3 employ Hilbert ordering instead~\cite{Wang2023OctFormer,wu2024ptv3}. 

However, these methods fail to capture causal, parent-child relationships between branches and can lead to disconnected or physically implausible structures.

Surprisingly, we found significantly better ordering strategies in the context of the trees. We first construct a tree graph representing the (botanical) tree's branching structure, where each node corresponds to a branch  $\mathbf{b}_i$ and edges capture parent-child relationships. We then traverse the tree using depth-first search (DFS) or breadth-first search (BFS). These traversal strategies determine the order of branches in the token sequence, ensuring that the hierarchical relationships within the tree are properly encoded. With these algorithms, we obtain a flattened sequence $\boldsymbol{\tau}_{\pi} = \pi(\boldsymbol{\tau})$ where $\pi(\cdot)$ represents a permutation applied to the original set.

This approach aligns with the tree's inherent structure and facilitates more coherent generation results by preserving the dependencies among branches. Additionally, these traversal methods ensure that each branch can inherit information from its ancestor nodes, further enhancing the consistency of the generation process.

\subsection{Hourglass Transformer Architecture}

\label{sec:hourglass_transformer}

Given the structured token sequence, we adopt an Hourglass Transformer~\cite{nawrot2021hierarchical,hao2024meshtron}  architecture to efficiently model long-range dependencies and hierarchical structure inherent in tree data.
\subsubsection{Architecture Overview}

As shown in Fig.~\ref{backbone}, the Hourglass Transformer employs a U-Net-like architecture, comprising distinct downsampling and upsampling stages interconnected via shift skip connections. This hierarchical design enables the model to effectively capture both global and local patterns within the input data, making it particularly suitable for tasks such as tree generation.

\subsubsection{Downsampling and Self-Attention}
Our network is specifically designed for structured trees with inherent structural. To aggregate this information progressively, the model utilizes a hierarchical downsampling strategy (local average pooling):

\begin{itemize}[leftmargin=*]
    \item \textbf{First Downsampling}: Reduces the sequence length by a factor of four (merging a quadruple of x, y, z, and r), resulting in each token representing one vertex of a branch.
    \item \textbf{Second Downsampling}: Further decreases the sequence length by a factor of two (merging two vertices), with each token corresponding to an entire branch.
\end{itemize}

At each downsampling scale, \emph{only one} self-attention layer is applied to model dependencies among tokens. Note that this differs from other U-Net style transformers, which still apply several layers in the outer stages.

\subsubsection{Bottleneck}

The bottleneck component serves as an intermediary between the downsampling and upsampling stages.
We build a mini-UNet-style bottleneck in the bottleneck which is proved to be effective in the experiments. The latter half of the bottleneck layers are connected to the first half, like a UNet.
In addition to the skip connections, we assign scales (learned during network training) to the first half of the layers,

\begin{equation}
b^i \leftarrow b^i + \alpha \circ,
\end{equation}
where $\alpha$ is the learned scale parameter, and $\circ$ represents the features from the first half of the layers.



This mechanism ensures that features captured in the earlier stages are adaptively reintroduced during the later stages, allowing the model to effectively utilize information from both ends of the bottleneck for more refined transformations.

\subsubsection{Upsampling and and Self-Attention}
In the decoder part, each compressed token is first upsampled (duplicated) to multiple ones. For example, the feature for branch $b^i$ is duplicated to $b^i_s$ and $b^i_t$ which represent the features for the two end-points of the branch $b^i$. Similarily for $b^i_s$ which is duplicated to $(b^i_{s, x}, b^i_{s, y}, b^i_{s, z}, b^i_{s, r})$. We perform \emph{only one} causal self-attention layer and connect it with the (shifted) outputs from the encoder side:
\begin{equation}
    \begin{aligned}
        b^i_s & \leftarrow \mathrm{CausalSA}(\mathrm{Upsample}(b^i_s)  + \mathrm{Shift}(b^i_s, \mathrm{Encoder})) \\
        b^i_t & \leftarrow \mathrm{CausalSA}(\mathrm{Upsample}(b^i_t)  + \mathrm{Shift}(b^i_t, \mathrm{Encoder})) \\
    \end{aligned}
\end{equation}
$\mathrm{Shift}(\cdot, \mathrm{Encoder})$ represents the element in the Encoder's output sequence shifted left by one.
We perform a similar operation for the sequence $((b^i_{s, x}, b^i_{s, y}, b^i_{s, z}, b^i_{s, r}), (b^i_{t, x}, b^i_{t, y}, b^i_{t, z}, b^i_{t, r}))$.

\subsubsection{Training Objective}

We train the autoregressive transformer using the standard cross-entropy loss. Given a training dataset of tokenized trees, the loss function is as follows.

\begin{equation}
\begin{aligned}
    \mathcal{L} &= -\sum_{j=1}^{8n} \log P(\square_i \mid \square_{<i}; \theta),
\end{aligned}
\end{equation}
where $8n$ is the total sequence length, $\square_i$ is the $i$-th token in the sequence, $\square_{<i}$ represents all the tokens before $\square_i$, and $\theta$ denotes the model parameters.


\subsubsection{Inference}

During sampling, the sequence length must be a multiplier of 8. To handle input sequences that do not perfectly align with the downsampling factors, the model includes padding operations to adjust the sequence length to the nearest multiple required by the downsampling layers. 

\section{Experiments}


For all experiments, unless otherwise stated, we employ data augmentation techniques, including random rotation along the z-axis and mirroring. The total batch size is set to 64, and the model is optimized using the AdamW optimizer with a learning rate of $5 \times 10^{-3}$. A warm-up period of 5 epochs is followed by a CosineAnnealingLR learning rate decay schedule. Flash Attention 2 is utilized for the attention mechanism. The training is conducted on 4 NVIDIA A100 GPUs (80GB each). 

All experiments utilized a transformer architecture with 24 layers. The first two downsampling layers each consist of one layer, followed by a bottleneck of 20 layers. The final two upsampling layers each comprised one layer. The transformer has 79,398,144 parameters, with an embedding dimension of 512 and 16 attention heads. For the Vision Transformer (ViT) encoder, we used the CLIP ViT from the timm library~\cite{rw2019timm}.

The training dataset was generated using the provided source code of the botanical tree growth model presented in \cite{Li:2023:Rhizomorph}.
Their package also provides realistic growth parameters for various tree species (represented as a 70-dimensional vector for each species), from which we selected the 6 species used in this paper.

For elm trees, we processed the dataset to limit each tree to a maximum of 200 branches and trained the model for 20 epochs. For other tree species, we limited each tree to up to 1,000 branches and trained for 30 epochs. In unconditional generation experiments, we utilized a dataset of 10,000 samples, divided into 9,000 for training and 1,000 for testing. In the conditional generation experiments, we employed 50,000 samples for training and 1,000 for testing, training for 50 epochs. In the Tree Growth Dynamics experiments, we generated sequences with a length of \(1027 \times 8\), training for 50 epochs. on 19,118 samples and testing on 1,000 samples.

\subsection{Evaluation Metrics}

To comprehensively assess the performance of our generative model, we employ a set of evaluation metrics that capture various aspects of the generated samples, including their novelty, uniqueness, quality, and similarity to real data distributions. The metrics are as follows:

\begin{itemize}[leftmargin=*]
    \item \textbf{FID (Fréchet Inception Distance)}: Measures the distance between the generated and real data distribution; lower values indicate higher fidelity.
    \item \textbf{PPL (Perplexity)}: Evaluates the stability and predictability of the generated sequences; lower values signify more coherent and stable generated results.
    \item \textbf{Connect}: Assesses the connectivity of the generated structures; higher values denote better structural coherence.
    \item \textbf{Novel}: Indicates the proportion of generated samples that are novel; a value of 1.0000 implies all samples are novel.
    \item \textbf{Unique}: Measures the uniqueness of generated samples; a value of 1.0000 implies all samples are unique.
    \item \textbf{MMD-CD (Maximum Mean Discrepancy - Chamfer Distance)}: Quantifies the similarity between generated and real data distributions using Chamfer Distance; lower values indicate closer alignment.
    \item \textbf{COV-CD (Coverage - Chamfer Distance)}: Measures the coverage of the real data distribution by the generated samples using Chamfer Distance; higher values denote better coverage.
    \item \textbf{JSD (Jensen-Shannon Divergence)}: Evaluates the similarity between the generated and real data distributions; lower values indicate higher similarity.
     \item \textbf{IoU (Intersection over Union)}: Measures the overlap between the generated image and the ground truth; higher values indicate better alignment.
      \item \textbf{Precision}: Assesses the accuracy of the generated structures in terms of correctly predicted elements; higher values indicate fewer false positives.
    \item \textbf{Recall}: Evaluates the ability of the model to capture all relevant elements of the real data; higher values indicate fewer false negatives.
    \item \textbf{F1-Score}: Combines Precision and Recall into a single metric by taking their harmonic mean; higher values indicate a better balance between Precision and Recall.

\end{itemize}

\subsection{Unconditional Generation}

We present the results of unconditional generation experiments. We begin by exploring various token-ordering strategies.
An ablation study is then conducted to evaluate the impact of specific design components on performance. Finally, we demonstrate the model's capability to generate diverse species, as illustrated in Fig.~\ref{fig:different_species_generation}.

\subsubsection{Token Ordering Strategies}

In our experiment on token ordering strategies, we evaluated four distinct methods---\textit{zyx}, \textit{Hilbert}, \textit{DFS}, and \textit{BFS} on \textbf{Elm} trees---across a comprehensive set of metrics, as presented in Table~\ref{tab:token_order_results}. 
\begin{table}[!ht]
\centering
\caption{Performance of different token ordering strategies.}
\vspace{-10pt}
\begin{tabular}{ccccc}
\hline
\textbf{Metric} & \textbf{zyx} & \textbf{hilbert} & \textbf{dfs} & \textbf{bfs} \\ \hline
\rowcolor{tabcol!30}\textbf{FID Score$\downarrow$} & 36.30 & 83.36 & \textbf{5.64} & 10.21 \\
\textbf{Connect$\uparrow$}      & 0.6233 & 0.2290 & \textbf{0.9866} & 0.9466 \\
\rowcolor{tabcol!30}\textbf{Novel$\uparrow$}        & \textbf{1.0000} & \textbf{1.0000} & \textbf{1.0000} & \textbf{1.0000} \\
\textbf{Unique$\uparrow$}       & \textbf{1.0000} & \textbf{1.0000} & \textbf{1.0000} & \textbf{1.0000} \\
\rowcolor{tabcol!30}\textbf{MMD-CD$\downarrow$}     & 0.0291 & 0.0260 & \textbf{0.0242} & 0.0268 \\
\textbf{COV-CD$\uparrow$}       & 0.4440 & 0.3850 & \textbf{0.5180} & 0.4850 \\
\rowcolor{tabcol!30}\textbf{JSD$\downarrow$}        & 0.0607 & 0.1012 & 0.0406 & \textbf{0.0375} \\ \hline
\vspace{-10pt}
\end{tabular}
\label{tab:token_order_results}
\end{table}

The results demonstrate that the \textit{DFS} token ordering strategy achieves the best performance. It attains the lowest FID  and the highest connectivity, reflecting its strong generative quality and structural coherence. Furthermore, \textit{DFS} maintains perfect novelty and uniqueness, indicating that its generated samples remain both distinct and diverse. The \textit{BFS} strategy also performs well, with an FID of 10.21, connectivity of 0.9466, and similarly favorable coverage and MMD-CD metrics. 

\begin{figure}[
!ht]
  \centering
  \includegraphics[width=\linewidth]{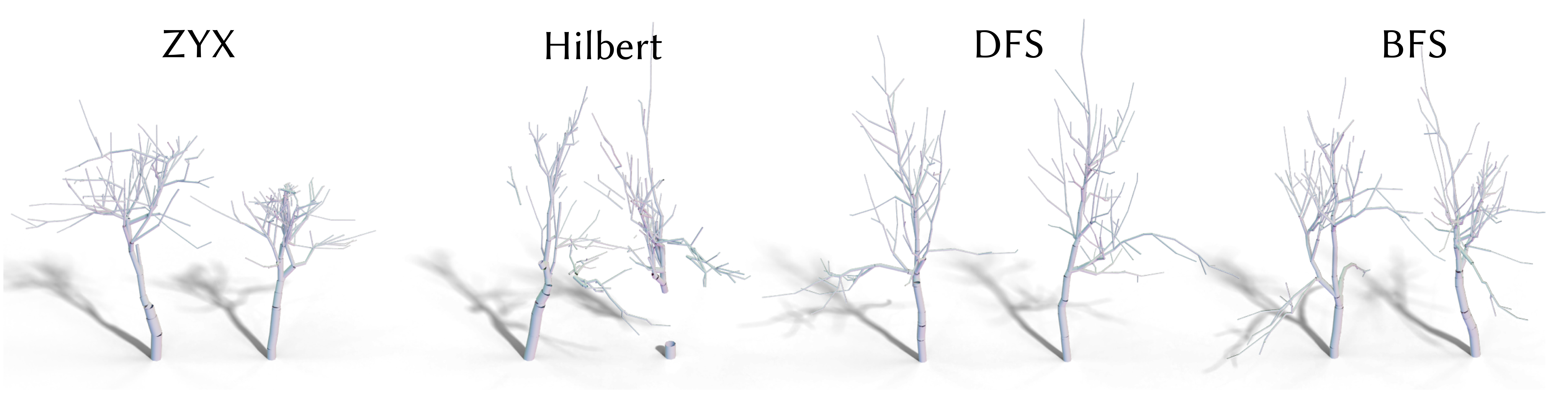}
  \vspace{-20pt}
  \caption{Tree structures generated using different token orders.}
  \vspace{-10pt}
  \Description{Comparison of token orders in tree generation.}
  \label{Fig:order}
\end{figure}

In contrast, both \textit{zyx} and \textit{Hilbert} exhibit substantially higher FID scores, suggesting lower generative quality. Their connectivity values (0.6233  and 0.2290, respectively) are also less favorable, although they retain perfect novelty and uniqueness. These findings confirm that while all ordering methods produce new and unique structures, the ability to capture the underlying data distribution and preserve coherent connections strongly depends on the chosen ordering strategy, as shown in Fig.~\ref{Fig:order}.

These results emphasize the critical impact of token ordering on generative performance. With \textit{DFS} and \textit{BFS} outperforming \textit{zyx} and \textit{Hilbert} across multiple evaluation metrics, we see clear evidence that a carefully designed token ordering strategy is essential for high-quality and diverse tree generation.

\subsubsection{Ablation Study}

To dissect the contributions of specific design components, an ablation study was conducted comparing five different model variants, as detailed in Table~\ref{tab:ablation_study}. The variants include:
\textbf{PT (Plain Transformer)} as the baseline architecture,
\textbf{HG1 (Hourglass 1)} incorporating an hourglass structure with one down-sampling layer,
\textbf{HG2 (Hourglass 2)} incorporating an hourglass structure with two down-sampling layers,
\textbf{HG2+R (Hourglass 2 with Bottleneck Residual)}, which adds non-learnable bottleneck residuals to HG2,
and \textbf{HG2+R+L (Hourglass 2 with Learnable Bottleneck Residual)} (or \textbf{Full}), which enhances HG2 with learnable bottleneck residuals.
\begin{table}[ht]
\centering
\caption{Performance comparison of different models. The best results in each metric are highlighted in bold.}
\vspace{-5pt}
\setlength{\tabcolsep}{4pt}
\begin{tabular}{l>{\columncolor{tabcol!30}}cc||>{\columncolor{tabcol!30}}cc>{\columncolor{tabcol!30}}c}
\hline
\textbf{Metric} & \textbf{PT} & \textbf{HG1} & \textbf{HG2} & \textbf{+R} & \textbf{+L (Full)} \\ \hline
\textbf{FID $\downarrow$}    & 9.111 & 9.263 & 6.186 & 5.996  & \textbf{5.641}         \\ 
\textbf{Time}                & 10m 15s & \textbf{5m 13s} & 5m 20s  & 5m 21s & 5m 21s        \\ 
\textbf{GPU}                 & 15.9G $\times$ 4 & \textbf{5.5G $\times$ 4} & 6.0G $\times$ 4 & 6.1G $\times$ 4 & 6.1G $\times$ 4 \\ \hline 
\end{tabular}
\vspace{-5pt}
\label{tab:ablation_study}
\end{table}

The baseline \textbf{PT} exhibited a high FID of 9.111, the longest training time of 10 minutes and 15 seconds, and the highest GPU memory usage at 15.9G $\times$ 4.
Introducing a single down-sampling layer in \textbf{HG1} decreased the training time to 5 minutes and 13 seconds, while GPU memory usage dropped to 5.5G $\times$ 4, but yields a similar FID with \textbf{PT}. Adding a second down-sampling layer in \textbf{HG2} substantially improved the FID to 6.186 with minimal changes to training time and GPU memory consumption.
The \textbf{HG2+R} variant, which incorporates non-learnable bottleneck residuals, achieved a reduction in FID to 5.996, with a slight increase in GPU usage to 6.1G $\times$ 4. Finally, the \textbf{Full} model, featuring learnable bottleneck residuals, attained the best FID of 5.641 while maintaining comparable training time and GPU memory usage to \textbf{HG2+R}.

\subsubsection{Scalability}

\begin{table}[h!]
\centering
\caption{Performance metrics of the model for different data sizes, showcasing scalability. Rows represent respective metrics with $\uparrow$ indicating higher is better and $\downarrow$ indicating lower is better.}
\vspace{-5pt}
\setlength{\tabcolsep}{4pt}
\begin{tabular}{lcccccc}
\hline
\textbf{Metric} & \textbf{9k} & \textbf{25k} & \textbf{50k} \\\hline
\rowcolor{tabcol!30}\textbf{FID Score$\downarrow$} & 5.6413 & 3.1792 & \textbf{2.9701} \\
\textbf{Connect$\uparrow$} & 0.9866 & 0.9983 & \textbf{0.9994} \\
\rowcolor{tabcol!30}\textbf{PPL$\downarrow$} & 0.8367 & 0.7847 & \textbf{0.7717} \\\hline
\end{tabular}
\vspace{-5pt}
\label{tab:model_scalability}
\end{table}

Table~\ref{tab:model_scalability} presents the evaluation metrics of the model for different dataset sizes: 9k, 25k, and 50k. These sizes correspond to varying amounts of training data, allowing us to assess the model's scalability.
The model shows strong scalability as the training data increases. The improvements in FID score, Connect, and PPL suggest that the model becomes more efficient and capable of generating high-quality, structurally sound, and diverse data as it is trained on larger datasets. These results emphasize the model’s potential for scaling to handle larger and more complex datasets effectively.

\subsubsection{Different species generation}

We present the results of our unconditional generation experiments. We trained 6 models for each species, including \textbf{Vitellaria }, \textbf{Hickory}, \textbf{Shadbush}, \textbf{Spruce}, and \textbf{Tulip}. As shown in Fig.~\ref{fig:different_species_generation}, our model is capable of generating diverse species with high fidelity. The generated samples exhibit significant variations in structure and appearance, demonstrating the model's capability to capture intricate patterns and characteristics inherent to different species.

\begin{table}[h!]
\centering
\caption{Performance metrics of different tree structures. Rows represent respective metrics with $\uparrow$ indicating higher is better and $\downarrow$ indicating lower is better.}
\label{tab:tree_metrics}
\vspace{-5pt}
\resizebox{0.98\linewidth}{!}{%
\setlength{\tabcolsep}{4pt}
\begin{tabular}{lccccc}
\hline
\textbf{Metric} & \textbf{Vitellaria } & \textbf{Hickory} & \textbf{Shadbush} & \textbf{Spruce} & \textbf{Tulip} \\\hline
\rowcolor{tabcol!30}\textbf{FID Score$\downarrow$} & 28.96 & 30.40 & 32.10 & 14.98 & 16.65 \\
\textbf{Connect$\uparrow$} & 0.9801 & 0.9880 & 0.9616 & 0.9825 & 0.9901 \\
\rowcolor{tabcol!30}\textbf{Novel$\uparrow$} & 1.0000 & 1.0000 & 1.0000 & 1.0000 & 1.0000 \\
\textbf{Unique$\uparrow$} & 1.0000 & 1.0000 & 1.0000 & 1.0000 & 1.0000 \\
\rowcolor{tabcol!30}\textbf{MMD-CD$\downarrow$} & 0.0390 & 0.0356 & 0.0247 & 0.0217 & 0.0048 \\
\textbf{COV-CD$\uparrow$} & 0.3060 & 0.3010 & 0.4420 & 0.3820 & 0.4300 \\
\rowcolor{tabcol!30}\textbf{JSD$\downarrow$} & 0.0959 & 0.1139 & 0.0976 & 0.0618 & 0.0378 \\\hline
\end{tabular}
}
\vspace{-5pt}
\end{table}

Table~\ref{tab:tree_metrics} summarizes the performance metrics of different tree structures generated by our model. Each row represents a specific evaluation metric. Notably, the model demonstrates high-quality results, with diverse and unique samples across all species, as indicated by the Unique and Novel metrics being consistently at their maximum value of 1.0. Additionally, the FID scores, which assess the fidelity of the generated samples, are relatively low across species, further suggesting the model's capability to generate high-fidelity tree structures. For the performance of the \textbf{Elm} species, please refer to Table~\ref{tab:token_order_results}.
\subsection{Modeling and Generating Tree Growth Dynamics}

In this section, we transform the ten distinct stages of tree growth into ten separate token lists, each representing a different development stage, and subsequently concatenate these lists in chronological order to form a single, continuous token sequence. As shown in Fig.~\ref{growing}, by employing an autoregressive generation approach, the model systematically learns and reconstructs the tree’s dynamic evolution from its initial sprout to full maturity. Concatenating the stage-wise token lists in this manner explicitly models the transitions between consecutive stages, thereby capturing critical inter-stage relationships that drive the tree’s morphological development over time. Moreover, the autoregressive framework leverages the entirety of the preceding stages to maintain and refine the details of the growth process. Consequently, the final generated sequence comprehensively depicts the tree’s complete growth trajectory.

\subsection{Conditional Generation}
\subsubsection{Completion}
Our model enables us to take a partial tree structure as an autoregressive prompt and continue generating the tree based on this initial input. As illustrated in Fig.~\ref{fig:treecompletion}, given an incomplete tree, our model can extend it into a diverse set of complete tree structures.


\subsubsection{Image to tree}

\begin{table}[ht]
\centering
\caption{Comparison of DFS and BFS ordering.}
\vspace{-5pt}
\begin{tabular}{lcccc}
\hline
\textbf{Metric} & \textbf{IoU$\uparrow$} & \textbf{Precision$\uparrow$} & \textbf{Recall$\uparrow$} & \textbf{F1-Score$\uparrow$} \\ \hline
\textbf{BFS}    & 0.2657       & 0.4038           & 0.4295        & 0.4153           \\ 
\rowcolor{tabcol!30}\textbf{DFS}    & \textbf{0.3045}     & \textbf{0.4332}            & \textbf{0.4958}          & \textbf{0.4615}       \\ \hline
\end{tabular}
\vspace{-5pt}
\label{tab:metric_comparison}
\end{table}

We explore the performance differences between Depth-First Search (DFS) and Breadth-First Search (BFS) strategies in the \textit{image2trees} task. As illustrated in Table~\ref{tab:metric_comparison}, DFS outperforms BFS across all evaluation metrics. Specifically, DFS achieves an Intersection over Union (IoU) of 0.3045, which is significantly higher than BFS's 0.2657. Precision and Recall are improved by approximately 7\% and 15\%, respectively, while the F1-Score increases by about 11\%. These results indicate that DFS is more effective in capturing key information from images when generating tree structures, thereby enhancing the overall quality of the generated results.

\begin{figure}[t]
  \centering
  \includegraphics[width=\linewidth]{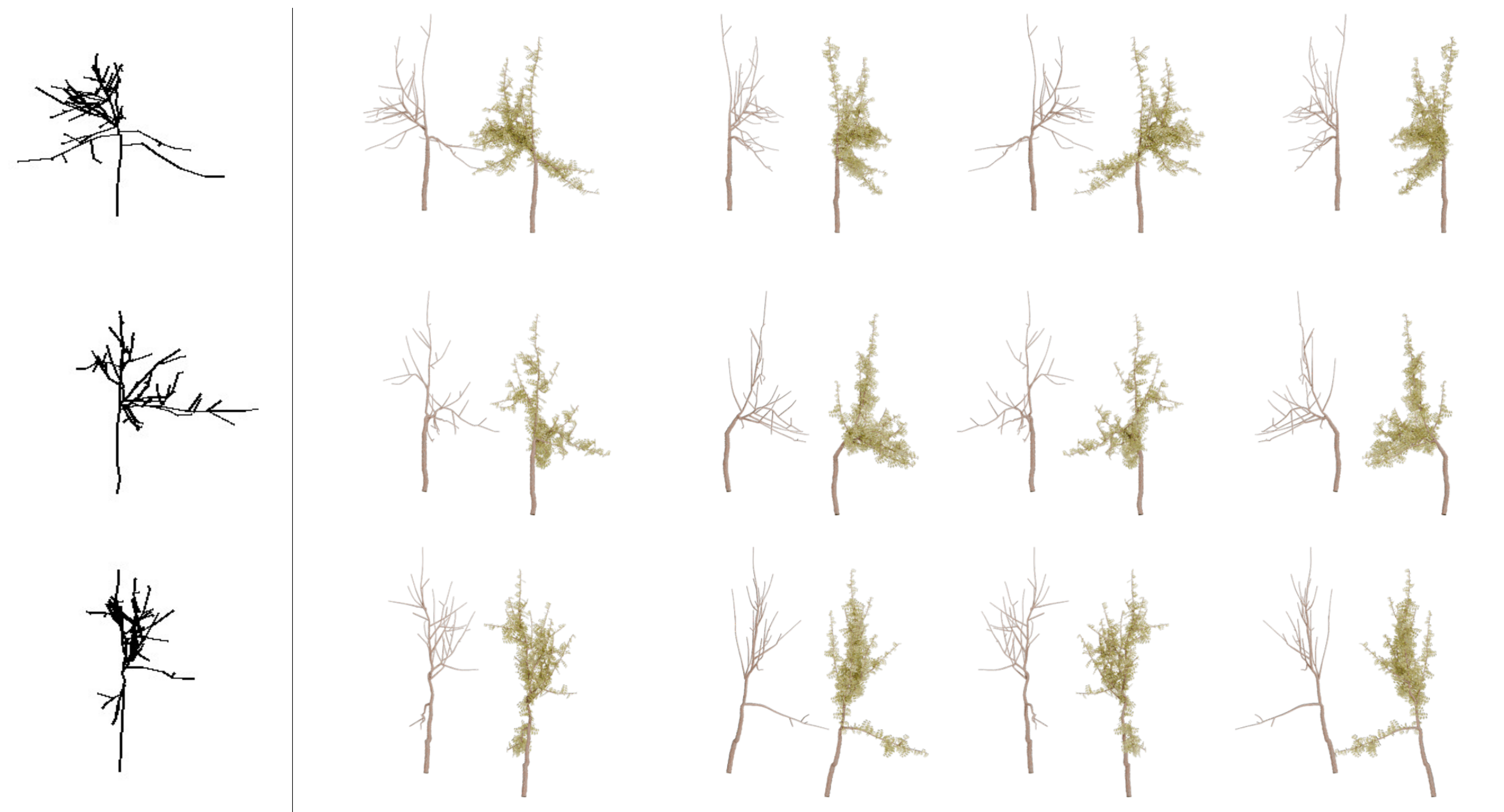}
  \caption{Tree structures generated using DFS token order. The first column illustrates the input image, while the subsequent four columns display different views of the resulting tree structures.}
  \label{fig:image2trees1}
\end{figure}

\begin{figure}[t]
  \centering
  \includegraphics[width=\linewidth]{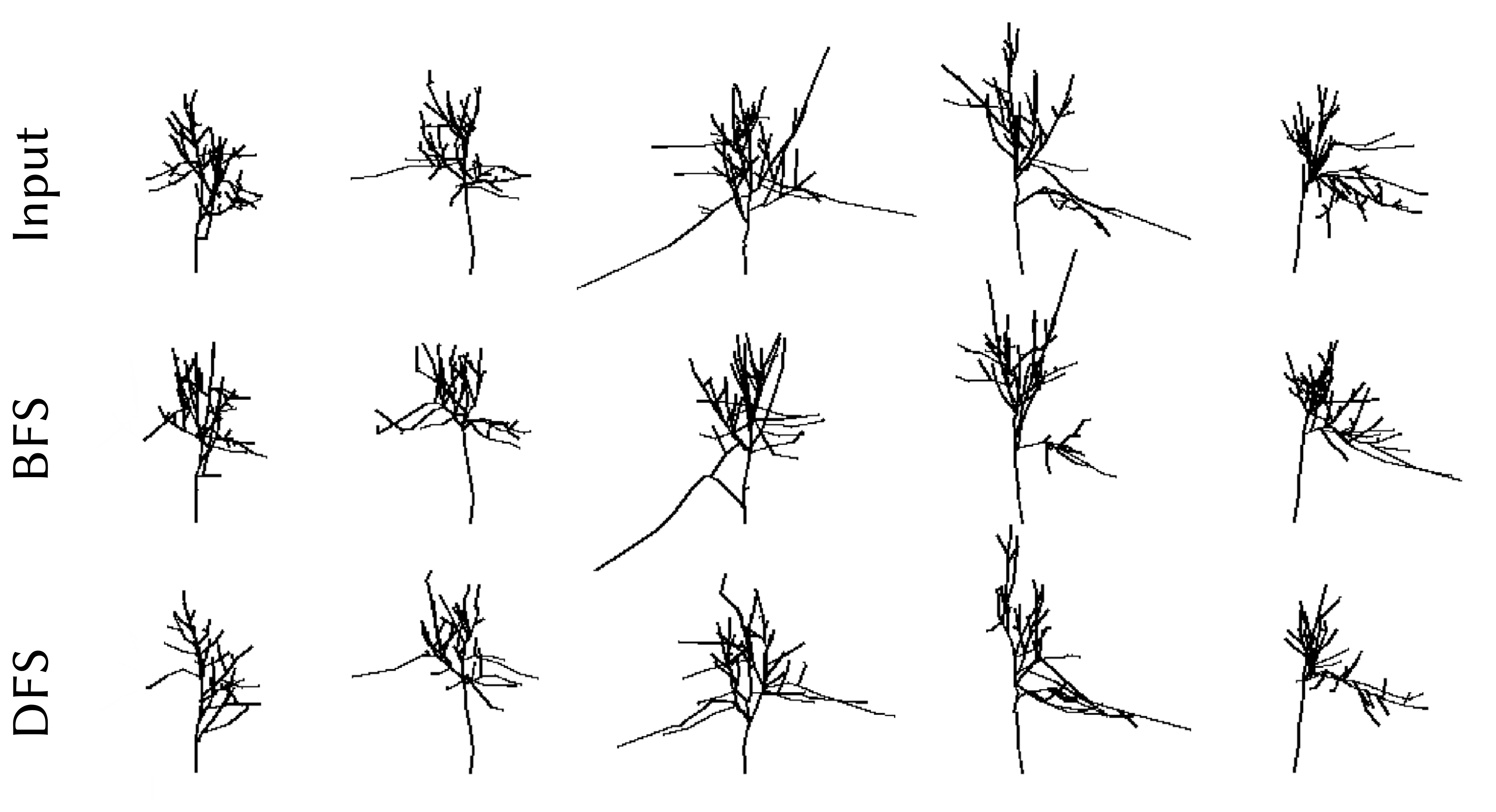}
  \caption{Comparison of tree structures generated using DFS and BFS token order.}
  \Description{Illustrates examples of tree structures generated by DFS and BFS token order, showing that DFS produces more detailed and accurate tree structures.}
  \label{fig:image2trees}
\end{figure}

Fig.~\ref{fig:image2trees} presents examples of tree structures generated using DFS and BFS. The trees generated by DFS exhibit more detailed and accurate node distributions with a deeper hierarchical organization. In contrast, those generated by BFS tend to be more balanced but may fail to capture finer details. This further validates the effectiveness of DFS in this task. Fig.~\ref{fig:image2trees1} shows different views of the resulting tree structures using DFS token order.

\begin{figure}[t]
  \centering
  
  \includegraphics[width=\linewidth]{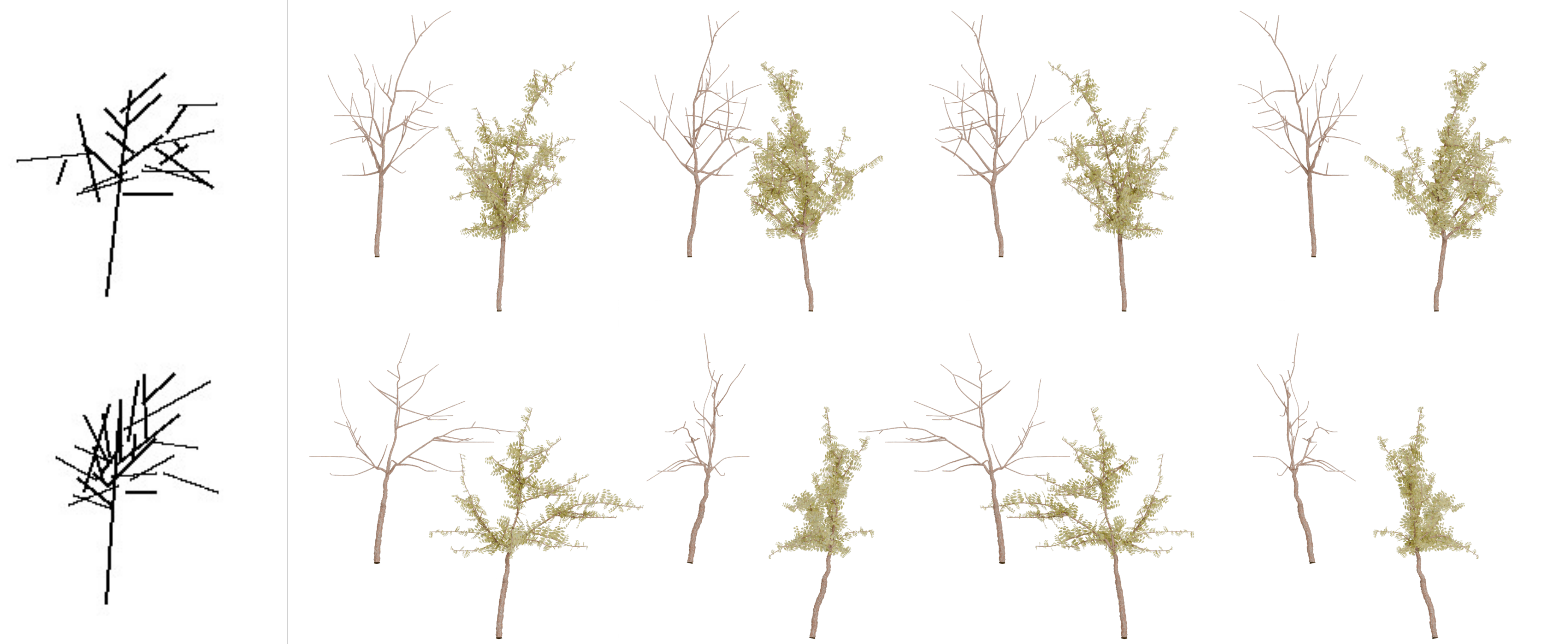}
  \caption{Tree structures conditioned on 2D vector graphics.}
  \label{fig:image2treessketch}
\end{figure}

Using our framework for training, we can also generate reasonable trees conditioned on hand-drawn 2D vector graphics, as shown in Fig.~\ref{fig:image2treessketch}. We successfully satisfy the intent of the interactive sketch in the main structure while generating diverse and plausible 3D branches, even though the input 2D vectors in the picture are not connected.

To achieve interactive tree modeling, we trained a conditional generative model using 2D vector graphics. Specifically, we projected the two endpoints of tree branches onto a 2D plane and rendered them as line segments, resulting in 2D vector graphics. 



\subsubsection{Point cloud to tree}

To achieve the generation of tree structures from sparse point cloud data, we randomly sample 200 points for each tree as the condition. We employ an approach that leverages 50 learnable query vectors and point cloud embeddings within a cross-attention framework.
This cross-attention mechanism effectively integrates the geometric information of the input point cloud into a fixed-length token list.

\begin{figure}[t]
  \centering
  \vspace{5pt}
  \includegraphics[width=\linewidth]{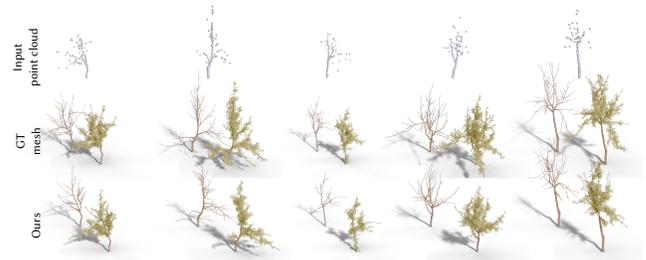}
  \caption{Point cloud to tree examples.}
  \Description{This figure illustrates generating tree structures from point cloud data using a cross-attention mechanism.}
  \label{fig:pc}
\end{figure}

Once the point cloud information is infused into the token list, it is utilized as a prompt for our tree autoregressive generation model. This model sequentially generates the tree structure, guided by the constraints and features extracted from the input point cloud. As depicted in Fig.~\ref{fig:pc}, our experimental results demonstrate that the generated tree models adhere closely to the input point cloud constraints, ensuring the overall structure aligns with the provided geometric data. 


\section{Conclusion and Future Work}

We introduced \textit{HourglassTree}, an innovative transformer-based architecture specifically engineered to generate 3D tree structures. We propose an efficient data structure, an hourglass-shaped transformer architecture, and suitable token ordering strategies for efficiently managing hierarchical tree data. 
Our framework extends to 4D growing trees and conditional generation tasks, including tree completion, image-to-tree, and point-cloud-to-tree generation.

Future work will explore several aspects to further enhance \textit{HourglassTree}. First, we aim to expand the scope of our approach to encompass a broader range of applications beyond trees, such as generating other organic or hierarchical structures (e.g., vascular systems, urban layouts, or molecular assemblies). Additionally, we will focus on optimizing the architecture to handle increasingly complex and larger-scale tree structures, potentially incorporating hierarchical parallelism and advanced memory management techniques. Investigating the application of linear attention mechanisms in tree generation tasks also presents a promising direction for enhancing efficiency and scalability.

In conclusion, \textit{HourglassTree} establishes a robust foundation for the next generation of deep generative tree models. We anticipate this work will inspire further research and development for generating structured 3D and 4D data.
\balance
\bibliographystyle{ACM-Reference-Format}
\bibliography{ref}

\appendix
\newpage

\begin{figure*}[p]
  \centering
  \includegraphics[width=0.85\textwidth]{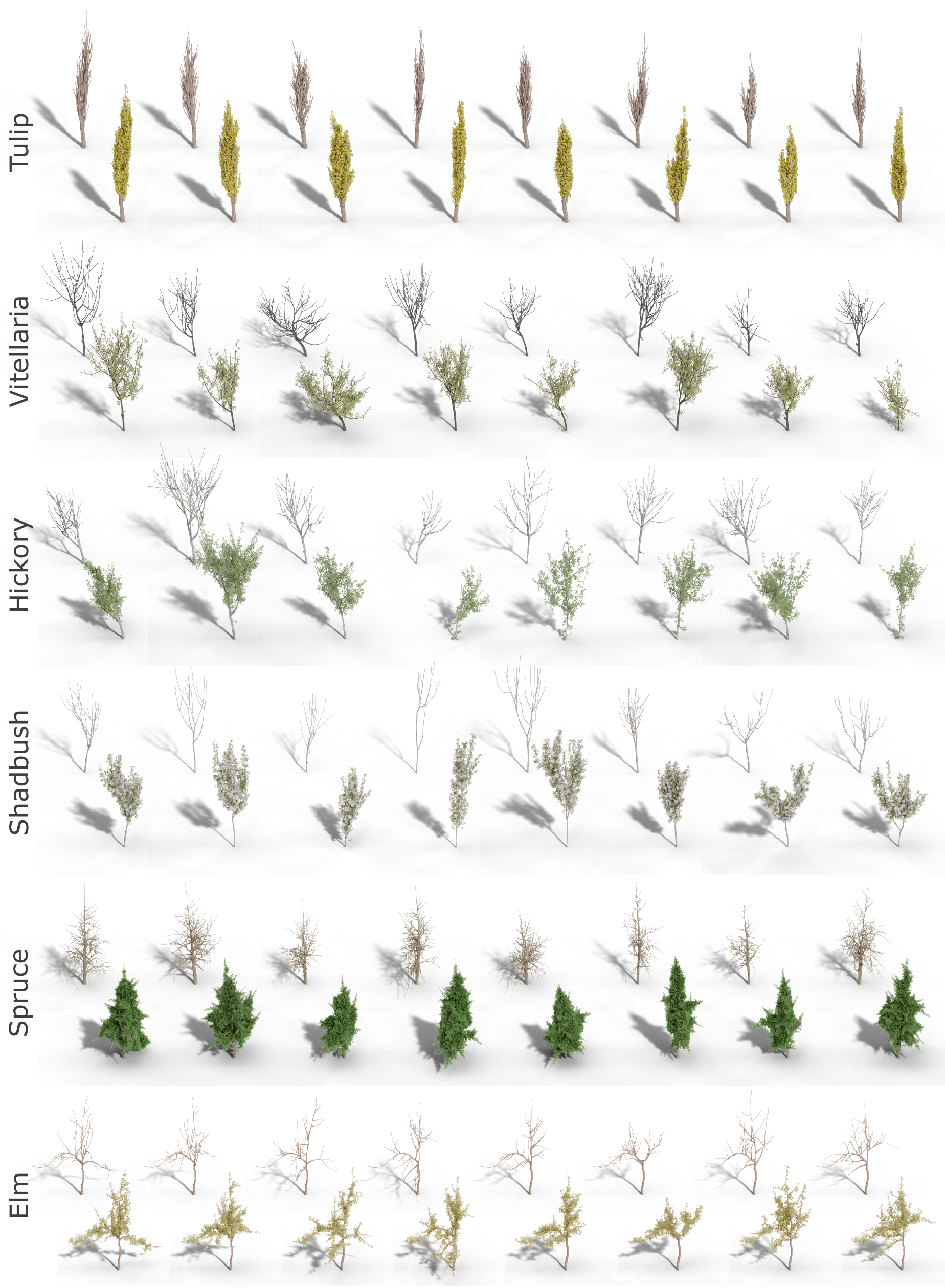}
  \caption{Unconditional generation of different tree species.}
  \label{fig:different_species_generation}
\end{figure*}

\begin{figure*}[h]
  \centering
  \includegraphics[width=0.85\textwidth]{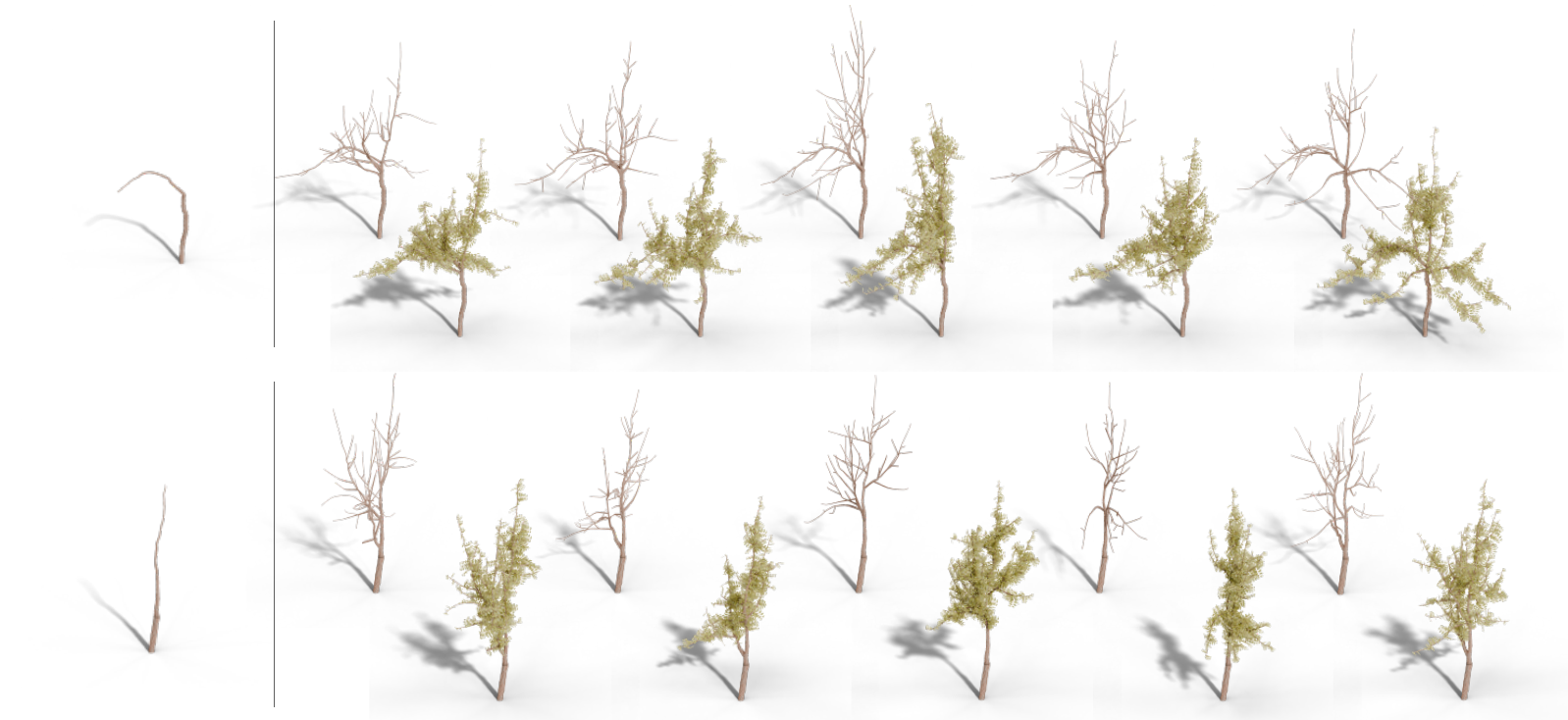}
  \caption{An illustration of tree completion. We show the input (left) and multiple different completions (right).}
  \Description{This figure illustrates the tree completion process, where a partial tree is provided as input, and the model generates a complete tree structure based on the initial prompt.}
  \label{fig:treecompletion}
\end{figure*}

\begin{figure*}
  \centering
  \includegraphics[width=0.85\textwidth]{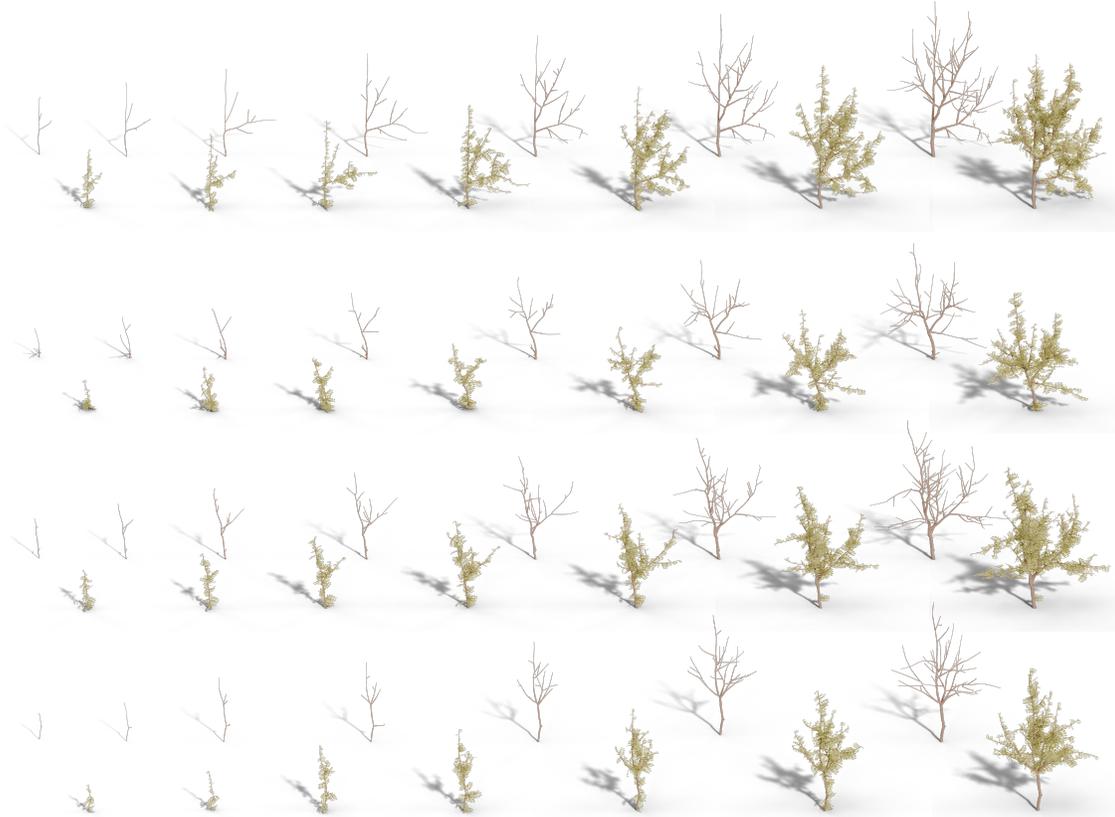}
  \caption{Unconditional generation of growing trees.}
  \Description{growing trees}
  \label{growing}
\end{figure*}


\end{document}